\documentclass[10pt,twocolumn,letterpaper]{article}

\usepackage{cvpr}
\usepackage{times}
\usepackage{epsfig}
\usepackage{graphicx}
\usepackage{amsmath}
\usepackage{amssymb}
\usepackage{booktabs}
\usepackage{caption} 
\usepackage{multirow}
\usepackage{wrapfig}
\usepackage{soul}
\usepackage{multirow}
\usepackage[normalem]{ulem}
\useunder{\uline}{\ul}{} 
\usepackage{changepage}
\usepackage{mathtools}
\usepackage[thinlines]{easytable}
\usepackage{makecell}

\DeclareMathOperator{\arccosh}{arccosh}
\newcommand{\fig}[1]{Figure~\ref{fig:#1}}

\usepackage{xfrac}

\usepackage[pagebackref=true,breaklinks=true,letterpaper=true,colorlinks,bookmarks=false]{hyperref}
\usepackage{cleveref}

\cvprfinalcopy
\def\cvprPaperID{7736} 

\ifcvprfinal\fi
\begin{document}

\title{Hyperbolic Image Embeddings}

\author{\textbf{Valentin Khrulkov}\textsuperscript{1,4}\thanks{Equal contribution} \quad \textbf{Leyla Mirvakhabova}\textsuperscript{1}\footnotemark[1] \quad \textbf{Evgeniya Ustinova}\textsuperscript{1} \\
\textbf{Ivan Oseledets}\textsuperscript{1,2} \quad \textbf{Victor Lempitsky}\textsuperscript{1,3}\\
\\
Skolkovo Institute of Science and Technology (Skoltech), Moscow\textsuperscript{1}\\
Institute of Numerical Mathematics of the
Russian Academy of Sciences, Moscow\textsuperscript{2} \\
Samsung AI Center, Moscow\textsuperscript{3} \\
Yandex, Moscow\textsuperscript{4} \\
{\tt\small \{valentin.khrulkov,leyla.mirvakhabova,evgeniya.ustinova,i.oseledets,lempitsky\}@skoltech.ru}}

\maketitle
\begin{abstract}
Computer vision tasks such as image classification, image retrieval, and few-shot learning are currently dominated by Euclidean and spherical embeddings so that the final decisions about class belongings or the degree of similarity are made using linear hyperplanes, Euclidean distances, or spherical geodesic distances (cosine similarity). In this work, we demonstrate that in many practical scenarios, hyperbolic embeddings provide a better alternative.
\end{abstract}

\section{Introduction}
\begin{figure}[htb!]
    \centering
    \includegraphics[width=0.4\textwidth]{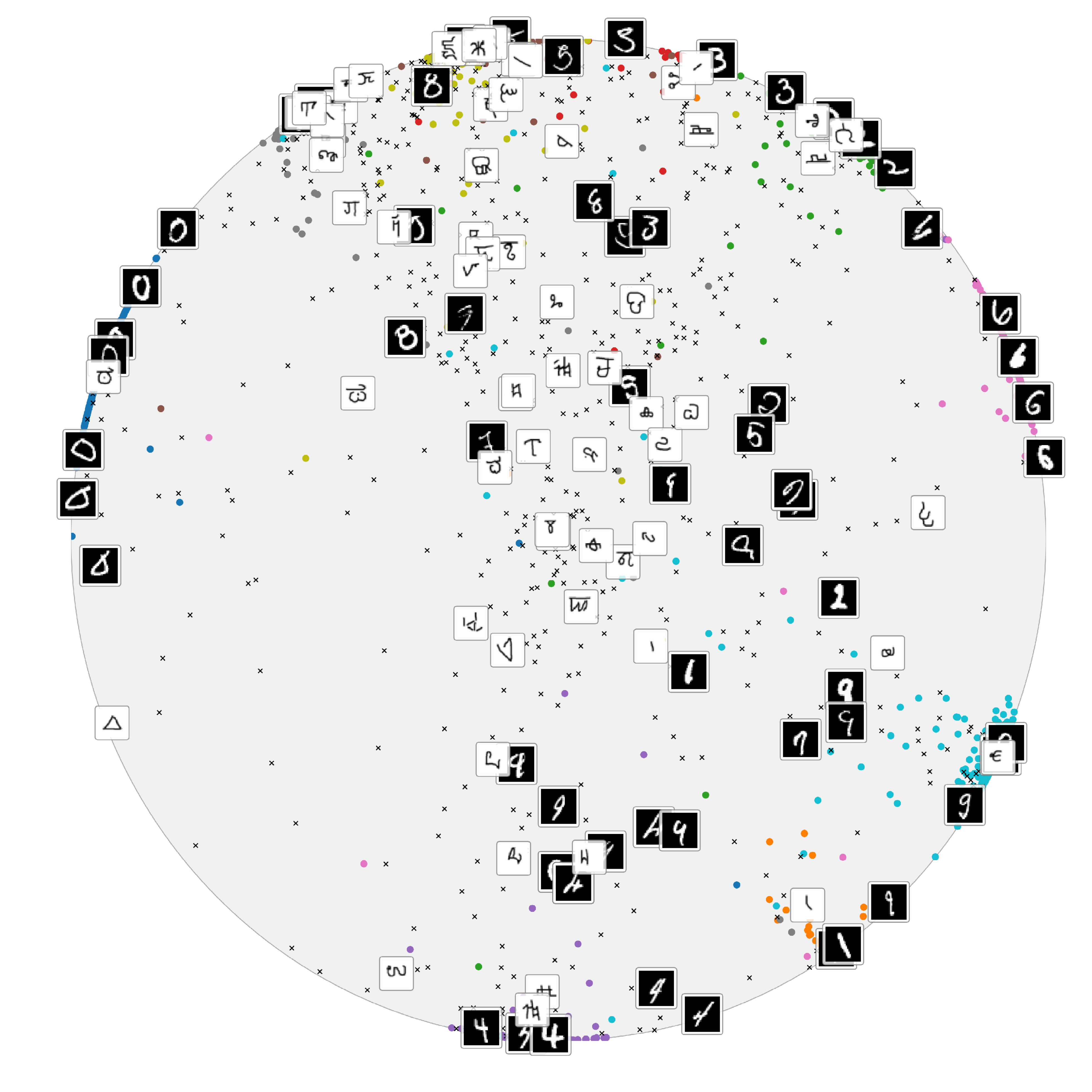}
    \caption{An example of two--dimensional Poincar\'e embeddings computed by a hyperbolic neural network trained on MNIST, and evaluated additionally on Omniglot. Ambiguous and unclear images from MNIST, as well as most of the images from Omniglot,  are embedded near the center, while samples with clear class labels (or characters from Omniglot similar to one of the digits) lie near the boundary. *For inference, Omniglot was normalized to have the same background color as MNIST. Omniglot images are marked with black crosses, MNIST images with colored dots.}
    \label{fig:teaser}
    \vspace{-5mm}
\end{figure}
Learned high-dimensional embeddings are ubiquitous in modern computer vision. Learning aims to group together semantically-similar images and to separate semantically-different images. When the learning process is successful, simple classifiers can be used to assign an image to classes, and simple distance measures can be used to assess the similarity between images or image fragments.
The operations at the end of deep networks imply a certain type of geometry of the embedding spaces. For example, image classification networks~\cite{krizhevsky2012imagenet, lecun1989generalization} use linear operators (matrix multiplication) to map embeddings in the penultimate layer to class logits. The class boundaries in the embedding space are thus piecewise-linear, and pairs of classes are separated by Euclidean hyperplanes. The embeddings learned by the model in the penultimate layer, therefore, live in the Euclidean space. The same can be said about systems where Euclidean distances are used to perform image retrieval~\cite{oh2016deep, sohn2016improved, wu2017sampling}, face recognition~\cite{Parkhi15, wen2016discriminative} or one-shot learning~\cite{snell2017prototypical_net}. 

Alternatively, some few-shot learning~\cite{nips2016_vinyals2016matching_net}, face recognition~\cite{schroff2015facenet}, and person re-identification methods~\cite{ustinova2016learning,Yi14} learn spherical embeddings, so that sphere projection operator is applied at the end of a network that computes the embeddings. Cosine similarity (closely associated with sphere geodesic distance) is then used by such architectures to match images.

Euclidean spaces with their zero curvature and spherical spaces with their positive curvature have certain profound implications on the nature of embeddings that existing computer vision systems can learn. In this work, we argue that hyperbolic spaces with negative curvature might often be more appropriate for learning embedding of images. Towards this end, we add the recently-proposed hyperbolic network layers~\cite{ganea2018hyperbolic} to the end of several computer vision networks, and present a number of experiments corresponding to image classification, one-shot, and few-shot learning and person re-identification. We show that in many cases, the use of hyperbolic geometry improves the performance over Euclidean or spherical embeddings. 

Our work is inspired by the recent body of works that demonstrate the advantage of learning hyperbolic embeddings for language entities such as taxonomy entries~\cite{nickel2017poincare}, common words~\cite{tifrea2018poincar}, phrases~\cite{dhingra2018embedding} and for other NLP tasks, such as neural machine translation ~\cite{gulcehre2018hyperbolic}.
Our results imply that hyperbolic spaces may be as valuable for improving the performance of computer vision systems.

\paragraph{Motivation for hyperbolic image embeddings.} 

\begin{figure}[t]
    \centering
    \includegraphics[width=\linewidth]{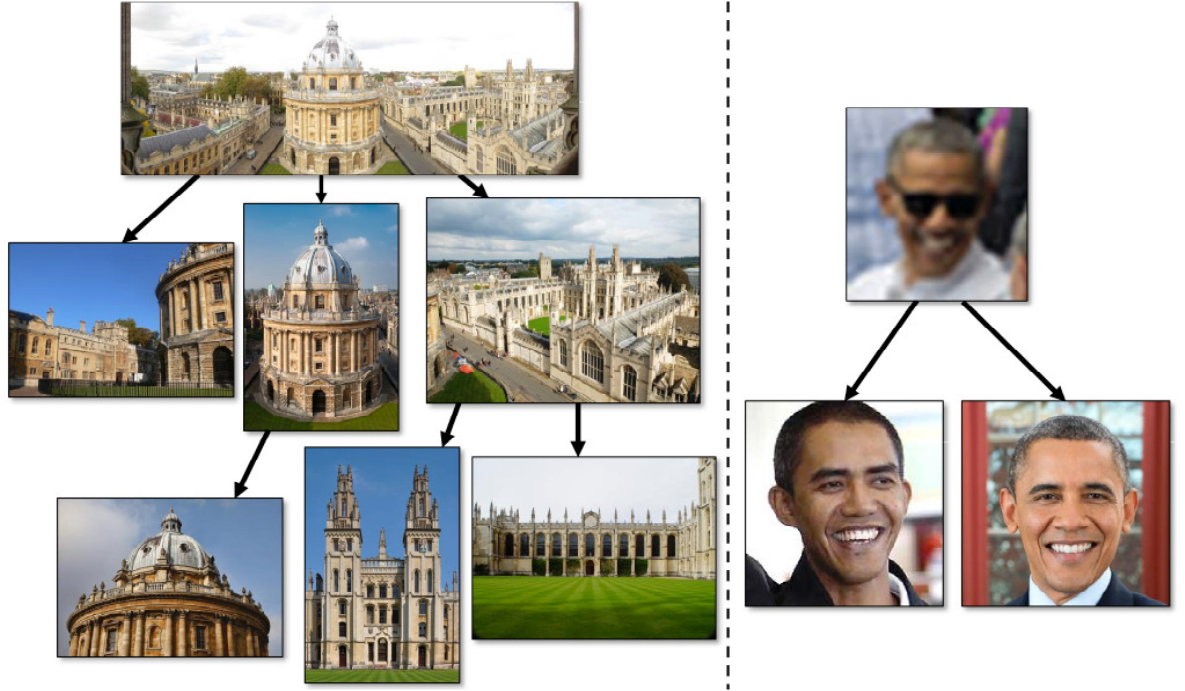}
    \caption{In many computer vision tasks, we want to learn image embeddings that obey the hierarchical constraints. E.g.,\ in image retrieval (left), the hierarchy may arise from whole-fragment relation. In  recognition tasks (right), the hierarchy can arise from image degradation, when degraded images are inherently ambiguous and may correspond to various identities/classes. Hyperbolic spaces are more suitable for embedding data with such hierarchical structure. }
    \label{fig:motivation}
    
\end{figure}

The use of hyperbolic spaces in natural language processing \cite{nickel2017poincare,tifrea2018poincar,dhingra2018embedding} is motivated by the ubiquity of hierarchies in NLP tasks. Hyperbolic spaces are naturally suited to embed hierarchies (e.g.,\ tree graphs) with low distortion ~\cite{sarkar2011low,sala2018representation}. Here, we argue that hierarchical relations between images are common in computer vision tasks (\fig{motivation}):
\begin{itemize}
\item In image retrieval, an overview photograph is related to many images that correspond to the close-ups of different distinct details. Likewise, for classification tasks in-the-wild, an image containing the representatives of multiple classes is related to images that contain representatives of the classes in isolation. Embedding a dataset that contains composite images into continuous space is, therefore, similar to embedding a hierarchy.
\item In some tasks, more generic images may correspond to images that contain less information and are therefore more ambiguous. E.g., in face recognition, a blurry and/or low-resolution face image taken from afar can be related to many high-resolution images of faces that clearly belong to distinct people. Again natural embeddings for image datasets that have widely varying image quality/ambiguity calls for retaining such hierarchical structure.
\item Many of the natural hierarchies investigated in natural language processing transcend to the visual domain. E.g.,\ the visual concepts of different animal species may be amenable for hierarchical grouping (e.g.~most felines share visual similarity while being visually distinct from pinnipeds).
\end{itemize}

Hierarchical relations between images call for the use of Hyperbolic spaces. Indeed, as the volume of hyperbolic spaces expands exponentially, it makes them continuous analogues of \emph{trees}, in contrast to Euclidean spaces, where the expansion is polynomial. It therefore seems plausible that the exponentially expanding hyperbolic space will be able to capture the underlying hierarchy of visual data. 

In order to build deep learning models which operate on the embeddings to hyperbolic spaces, we capitalize on recent developments ~\cite{ganea2018hyperbolic}, which construct the analogues of familiar layers (such as a feed--forward layer, or a multinomial regression layer) in hyperbolic spaces. We show that many standard architectures used for tasks of image classification, and in particular in the few--shot learning setting can be easily modified to operate on hyperbolic embeddings, which in many cases also leads to their improvement.

The main contributions of our paper are twofold:
\begin{itemize}
\item First, we apply the machinery of hyperbolic neural networks to computer vision tasks. Our experiments with various few-shot learning and person re-identification models and datasets demonstrate that hyperbolic embeddings are beneficial for visual data. 
\item Second, we propose an approach to evaluate the hyperbolicity of a dataset based on the concept of Gromov $\delta$-hyperbolicity. It further allows estimating the radius of Poincar\'e disk for an embedding of a specific dataset and thus can serve as a handy tool for practitioners.
\end{itemize}

\section{Related work}
\paragraph{Hyperbolic language embeddings.} Hyperbolic embeddings in the natural language processing field have recently been very successful~\cite{nickel2017poincare, Nickel18}. They are motivated by the innate ability of hyperbolic spaces to embed hierarchies (e.g., \ tree graphs) with low distortion~\cite{sala2018representation,sarkar2011low}. However, due to the discrete nature of data in NLP, such works typically employ Riemannian optimization algorithms in order to learn embeddings of individual words to hyperbolic space. This approach is difficult to extend to visual data, where image representations are typically computed using CNNs.

Another direction of research, more relevant to the present work, is based on imposing hyperbolic structure on \emph{activations of neural networks} \cite{ganea2018hyperbolic,gulcehre2018hyperbolic}. However, the proposed architectures were mostly evaluated on various NLP tasks, with correspondingly modified traditional models such as RNNs or Transformers. We find that certain computer vision problems that heavily use image embeddings can benefit from such hyperbolic architectures as well. Concretely, we analyze the following tasks.
\vspace{-5mm}
\paragraph{Few--shot learning.} 
The task of few--shot learning is concerned with the overall ability of the model to generalize to unseen data during training. Most of the existing state-of-the-art few--shot learning models are based on metric learning approaches, utilizing the distance between image representations computed by deep neural networks as a measure of similarity \cite{nips2016_vinyals2016matching_net,snell2017prototypical_net,sung2018relation_net,nichol2018reptile,chen2018a,chu_cvpr2019_spot_and_learn,li_cvpr2019_revisiting,bauer2017discriminative_k_shot,rusu2018LEO,chen2019aug_few_shot}. In contrast, other models apply meta-learning to few-shot learning: e.g.,\ MAML  by ~\cite{finn2017maml}, Meta-Learner LSTM by ~\cite{ravi2016optimization}, SNAIL by ~\cite{mishra2017snail}. While these methods employ either Euclidean or spherical geometries (like in ~\cite{nips2016_vinyals2016matching_net}), there was no extension to hyperbolic spaces. 
\vspace{-5mm}
\paragraph{Person re-identification.} 
The task of person re-identification is to match pedestrian images captured by possibly non-overlapping surveillance cameras. 
Papers \cite{ahmed2015improved, guo2018efficient, wang2018person} adopt the pairwise models that accept pairs of images and output their similarity scores. The resulting similarity scores are used to classify the input pairs as being matching or non-matching. 
Another popular direction of work includes approaches that aim at learning a mapping of the pedestrian images to the Euclidean descriptor space. Several papers, e.g.,\ \cite{suh2018part,Yi14} use verification loss functions based on the Euclidean distance or cosine similarity.  
A number of methods utilize a simple classification approach for training \cite{chang2018multi, su2017pose,kalayeh2018human,zhao2017spindle}, and Euclidean distance is used in test time. 
\section{Reminder on hyperbolic spaces and hyperbolicity estimation.}\label{sec:pball}
Formally, $n$-dimensional hyperbolic space denoted as $\mathbb{H}^n$ is defined as the homogeneous, simply connected $n$-dimensional Riemannian manifold of constant negative sectional curvature. The property of constant negative curvature makes it analogous to the ordinary Euclidean sphere (which has constant positive curvature); however, the geometrical properties of the hyperbolic space are very different. It is known that hyperbolic space cannot be isometrically embedded into Euclidean space ~\cite{krioukov2010hyperbolic,linial1998low}, but there exist several well--studied \emph{models} of hyperbolic geometry. In every model, a certain subset of Euclidean space is endowed with a \emph{hyperbolic metric}; however, all these models are isomorphic to each other, and we may easily move from one to another base on where the formulas of interest are easier. We follow the majority of NLP works and use the \textit{Poincar\'e ball} model. 

The Poincar\'e ball model $(\mathbb{D}^n, g^{\mathbb{D}})$ is defined by the manifold $\mathbb{D}^n = \lbrace \mathbf{x} \in \mathbb{R}^n \colon \|\mathbf{x}\| < 1 \rbrace$ endowed with the Riemannian metric $g^{\mathbb{D}}(\mathbf{x})=\lambda_{\mathbf{x}}^2 g^{E}$, where $\lambda_{\mathbf{x}} = \frac{2}{1 - \|\mathbf{x}\|^2}$ is the \emph{conformal factor} and $g^{E}$ is the Euclidean metric tensor $g^{E} = \mathbf{I}^n$. In this model the \emph{geodesic distance} between two points is given by the following expression:
\begin{equation}\label{dist}
    d_{\mathbb{D}}(\mathbf{x}, \mathbf{y}) = \arccosh\Big(1 + 2 \frac{\|\mathbf{x} - \mathbf{y}\|^2}{(1 - \|\mathbf{x}\|^2)(1-\|\mathbf{y}\|^2)}\Big).
\end{equation}

\begin{figure}[htb!]
    \centering
    \includegraphics[width=0.8\linewidth]{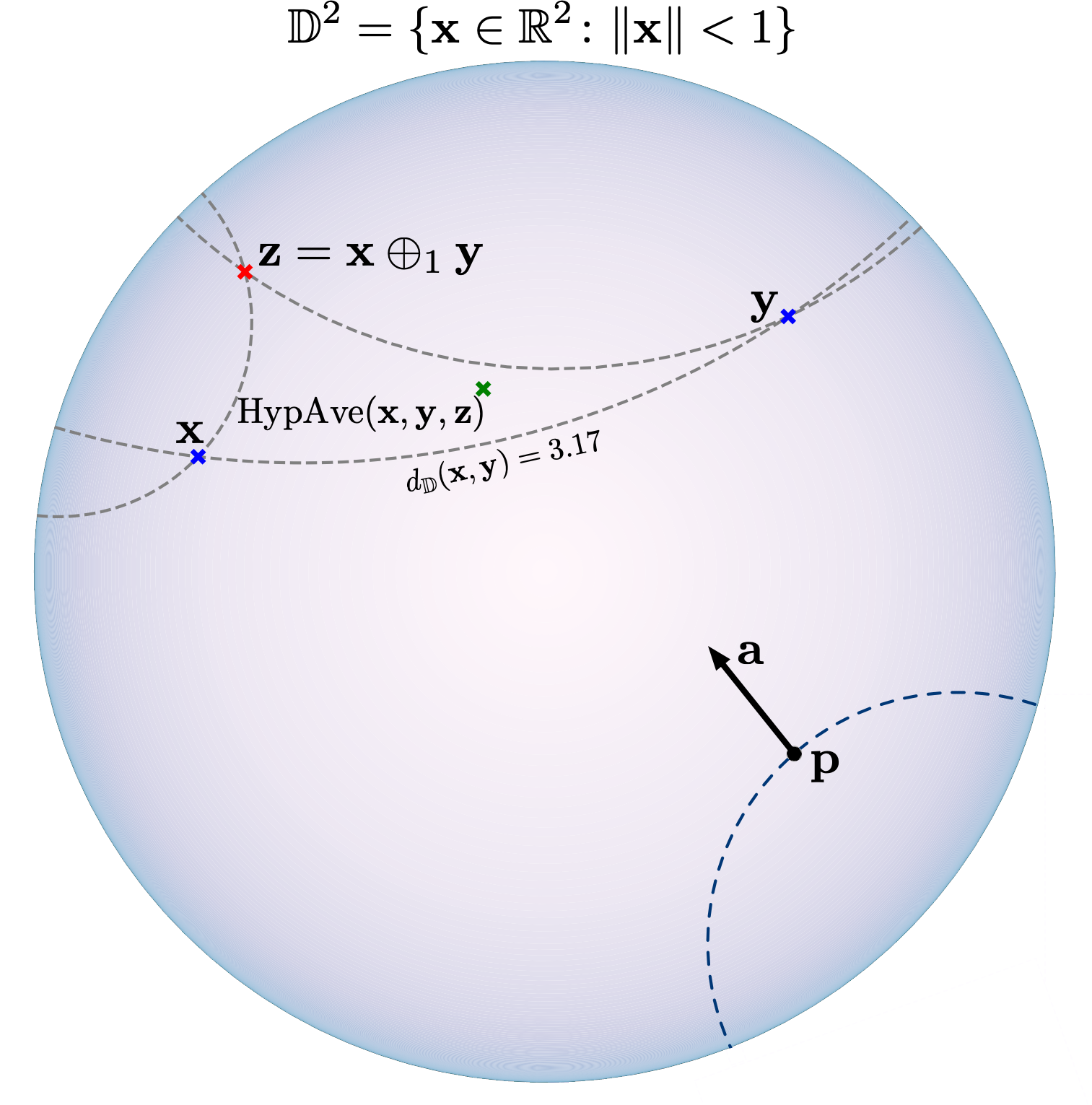}
    \caption{Visualization of the two--dimensional Poincar\'e ball. Point $\mathbf{z}$ represents the \emph{M\"obius sum} of points $\mathbf{x}$ and $\mathbf{y}$. $\mathrm{HypAve}$ stands for hyperbolic averaging. Gray lines represent \emph{geodesics}, curves of shortest length connecting two points. In order to specify the \emph{hyperbolic hyperplanes} (bottom), used for multiclass logistic regression, one has
    to provide an origin point $\mathbf{p}$ and a normal vector $\mathbf{a} \in T_{\mathbf{p}}\mathbb{D}^2 \setminus \{\mathbf{0}\}$.
    For more details on hyperbolic operations see Section \ref{sec:hyp-neural}.
    }
    \label{fig:pball_2d}
\end{figure}

In order to define the \emph{hyperbolic average}, we will make use of the \emph{Klein model} of hyperbolic space. Similarly to the Poincar\'e model, it is defined on the set \mbox{$\mathbb{K}^n = \lbrace \mathbf{x} \in \mathbb{R}^n : \|\mathbf{x}\| < 1 \rbrace$}, however, with a different metric, not relevant for further discussion. In Klein coordinates, the hyperbolic average (generalizing the usual Euclidean mean) takes the most simple form, and we present the necessary formulas in Section~\ref{sec:hyp-neural}.

From the viewpoint of hyperbolic geometry, all points of Poincar\'e ball are equivalent. The models that we consider below are, however, hybrid in the sense that most layers use Euclidean operators, such as standard generalized convolutions, while only the final layers operate within the hyperbolic geometry framework. The hybrid nature of our setups makes the origin a special point, since, from the Euclidean viewpoint, the local volumes in Poincare ball expand exponentially from the origin to the boundary. This leads to the useful tendency of the learned embeddings to place more generic/ambiguous objects closer to the origin while moving more specific objects towards the boundary. The distance to the origin in our models, therefore, provides a natural estimate of uncertainty, that can be used in several ways, as we show below.

This choice is justified for the following reasons. First, many existing vision architectures are designed to output embeddings in the vicinity of zero (e.g., in the unit ball).  
Another appealing property of hyperbolic space (assuming the standard Poincare ball model) is the existence of a reference point -- the center of the ball. We show that in image classification which construct embeddings in the Poincare model of hyperbolic spaces the distance to the center can serve as a measure of \emph{confidence} of the model --- the input images which are more familiar to the model get mapped closer to the boundary, and images which confuse the model (e.g., blurry or noisy images, instances of a previously unseen class) are mapped closer to the center.
The geometrical properties of hyperbolic spaces are quite different from the properties of the Euclidean space. For instance, the sum of angles of a geodesic triangle is always less than $\pi$. These interesting geometrical properties make it possible to construct a ``score'' which for an arbitrary metric space provides a degree of similarity of this metric space to a hyperbolic space. This score is called $\delta$-hyperbolicity, and we now discuss it in detail.

\subsection{$\delta$-Hyperbolicity}
\begin{figure}
    \centering
    \includegraphics[width=0.5\linewidth]{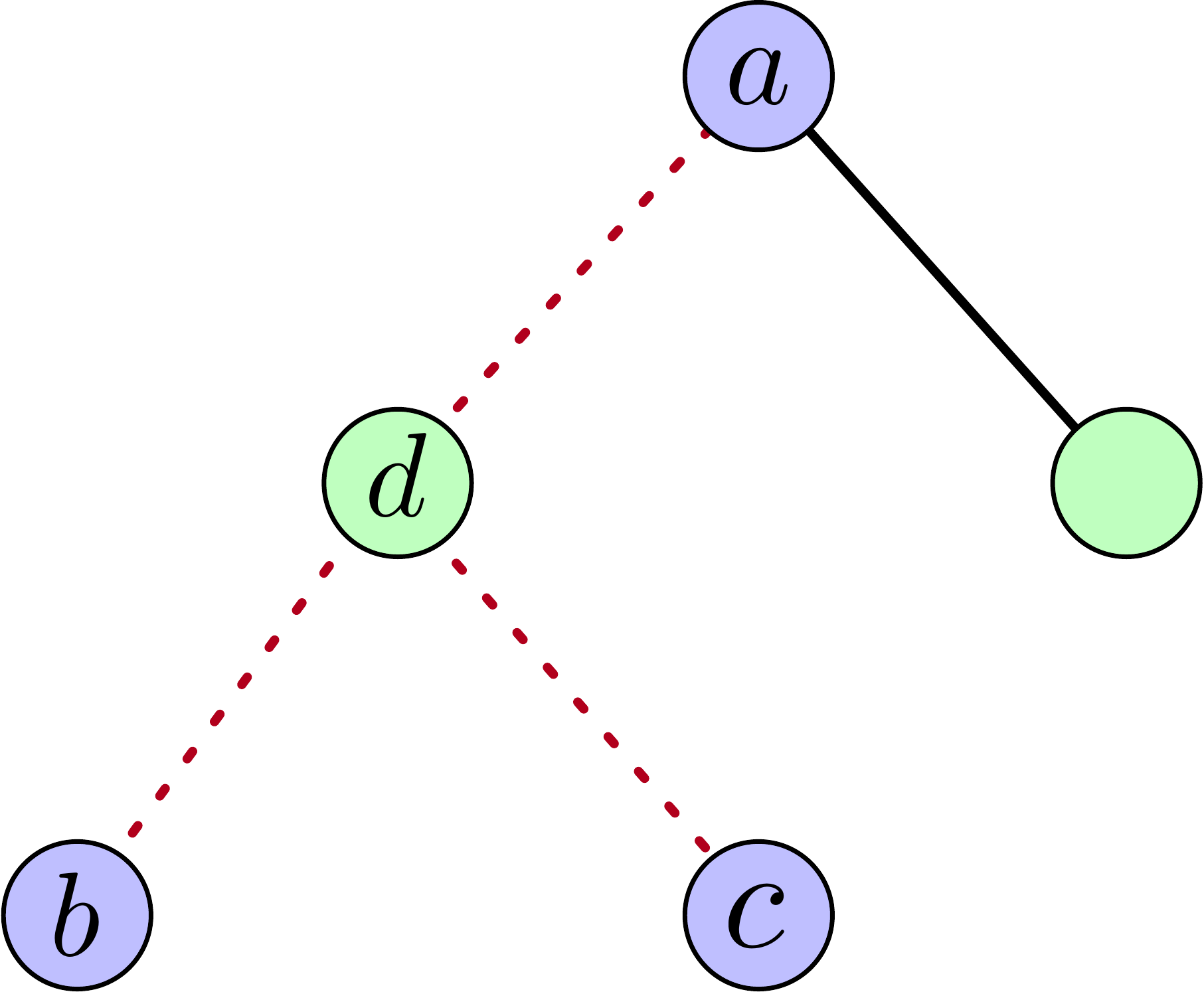}
    \caption{Visualization of a \emph{geodesic triangle} in a tree. Such a tree endowed with a natural shortest path metric is a $0$--Hyperbolic space.}
    \label{fig:tree}
    \vspace{-2mm}
\end{figure}

Let us start with an illustrative example. The simplest discrete metric space possessing hyperbolic properties is a \emph{tree} (in the sense of graph theory) endowed with the natural shortest path distance. Note the following property: for any three vertices $a, b, c$, the geodesic triangle (consisting of geodesics --- paths of shortest length connecting each pair) spanned by these vertices (see \Cref{fig:tree}) is \emph{slim}, which informally means that it has a center (vertex $d$) which is contained in every side of the triangle. By relaxing this condition to allow for some slack value $\delta$ and considering so-called $\delta$-slim triangles, we arrive at the following general definition. 

 \begin{table}[htb!]
\caption{Comparison of the theoretical degree of hyperbolicity with the relative delta $\delta_{rel}$ values estimated using \Cref{gromov_product,eq:min_max}. The numbers are given for the two-dimensional Poincar\'e ball $\mathbb{D}^2$, the 2D sphere $S_2$, the upper hemisphere $S_2^{+}$, and a (random) tree graph. 
}
\setlength{\tabcolsep}{4pt}
\label{tab:deltas_toy}
\vspace{-5mm}
\begin{center}
\begin{tabular}{lcccc}
\toprule
{} &  $\mathbb{D}^2$ &  $S_2^{+}$ &  $S_2$ &  Tree \\
\midrule
Theory &  $0$ & $1$ &    $1$ &   $0$ \\
$\delta_{rel}$  &  $0.18\pm0.08$ & $0.86 \pm 0.11$ &    $0.97\pm0.13$ & $0.0$ \\
\bottomrule
\end{tabular}
\end{center}
\vspace{-5mm}
\end{table}


\begin{table}[htb!]
\caption{The relative delta $\delta_{rel}$ values calculated for different datasets. For image datasets we measured the Euclidean distance between the features produced by various standard feature extractors pretrained on ImageNet. Values of $\delta_{rel}$ closer to $0$ indicate a stronger hyperbolicity of a dataset. Results are averaged across $10$ subsamples of size $1000$. The standard deviation for all the experiments did not exceed $0.02$.
}
\setlength{\tabcolsep}{4pt}
\label{tab:deltas_cv}
\vspace{-5mm}
\begin{center}
 \scalebox{0.9}{
\begin{tabular}{lcccc}
\toprule
\multirow{2}{*}{\textbf{Encoder}} &  \multicolumn{4}{c}{\textbf{Dataset}} \\

& CIFAR10 &  CIFAR100 &  CUB &   \emph{Mini}ImageNet \\
\midrule
 Inception v3 \cite{szegedy2015going}   & $0.25$ &  $0.23$ &  $0.23$ &  $0.21$ \\
ResNet34 \cite{he2016deep}    &  $0.26$ &   $0.25$ &   $0.25$ &  $0.21$ \\
VGG19 \cite{simonyan2014very} &  $0.23$ &   $0.22$ &  $0.23$ & $0.17$ \\
\bottomrule
\end{tabular}
}
\end{center}
\vspace{-5mm}
\end{table}

Let $X$ be an arbitrary (metric) space endowed with the distance function $d$. Its $\delta$-hyperbolicity value then may be computed as follows. We start with the so-called $\emph{Gromov product}$ for points $x,y,z \in X$:
\begin{equation}\label{gromov_product}
(y,z)_x= \frac{1}{2} (d(x,y) +d(x,z) - d(y,z)).
\end{equation}
Then, $\delta$ is defined as the minimal value such that the following four-point condition holds for all points $x,y,z,w \in X$:
\begin{equation}
(x,z)_w \geq \min((x,y)_w,(y,z)_w) - \delta.
\end{equation}
The definition of hyperbolic space in terms of the Gromov product can be seen as saying that the metric relations between any four points are the same as they would be in a tree, up to the additive constant $\delta$. $\delta$-Hyperbolicity captures the basic common features of ``negatively curved'' spaces like the classical
real-hyperbolic space $\mathbb{D}^n$ and of discrete spaces like trees.

For practical computations, it suffices to find the $\delta$ value for some fixed point $w = w_0$ as it is independent of $w$. An efficient way to compute $\delta$ is presented in~\cite{fournier2015computing}. Having a set of points, we first compute the matrix $A$ of pairwise Gromov products using \Cref{gromov_product}. After that, the $\delta$ value is simply the largest coefficient in the matrix $(A \otimes A) - A$, where $\otimes$ denotes the min-max matrix product 
\begin{equation}{\label{eq:min_max}}
    A \otimes B = \max_k \min \{A_{ik}, B_{kj}\}.
\end{equation}

\paragraph{Results.}
 In order to verify our hypothesis on hyperbolicity of visual datasets we compute the scale-invariant metric, defined as \mbox{$\delta_{rel}(X) = \frac{2\delta(X)}{\mathrm{diam}(X)}$}, where $\mathrm{diam}(X)$ denotes the set diameter (maximal pairwise distance). By construction, $\delta_{rel}(X) \in [0, 1]$ and specifies how close is a dataset to a hyperbolic space. Due to computational complexities of \Cref{gromov_product,eq:min_max} we employ the batched version of the algorithm, simply sampling $N$ points from a dataset, and finding the corresponding $\delta_{rel}$. Results are averaged across multiple runs, and we provide resulting mean and standard deviation. We experiment on a number of toy datasets (such as samples from the standard two--dimensional unit sphere), as well as on a number of popular computer vision datasets. As a natural distance between images, we used the standard Euclidean distance between feature vectors extracted by various CNNs pretrained on the ImageNet (ILSVRC) dataset~\cite{deng2009imagenet}. Specifically, we consider VGG19 \cite{simonyan2014very}, ResNet34 \cite{he2016deep} and Inception v3 \cite{szegedy2015going} networks for  distance evaluation. While other metrics are possible, we hypothesize that the underlying hierarchical structure (useful for computer vision tasks) of image datasets can be well understood in terms of their deep feature similarity.
 
 

Our results are summarized in \Cref{tab:deltas_cv}. We observe that the degree of hyperbolicity in image datasets is quite high, as the obtained $\delta_{rel}$ are significantly closer to $0$ than to $1$ (which would indicate complete non-hyperbolicity). This observation suggests that visual tasks can benefit from hyperbolic representations of images. 

\paragraph{Relation between $\delta$-hyperbolicity and Poincar\'e disk radius.}
It is known \cite{tifrea2018poincar} that the standard Poincar\'e ball is $\delta$-hyperbolic with $\delta_P = \log(1 + \sqrt{2}) \approx 0.88$. Formally, the diameter of the Poincar\'e ball is infinite, which yields the $\delta_{rel}$ value of $0$. However, from computational point of view we cannot approach the boundary infinitely close. Thus, we can compute the \emph{effective} value of $\delta_{rel}$ for the Poincar\'e ball. For the clipping value of $10^{-5}$, i.e., when we consider only the subset of points with the (Euclidean) norm not exceeding $1 - 10^{-5}$, the resulting diameter is equal to $\sim 12.204$. This provides the effective $\delta_{rel} \approx 0.144$. Using this constant we can estimate the radius of Poincar\'e disk suitable for an embedding of a specific dataset. Suppose that for some dataset $X$ we have found that its $\delta_{rel}$ is equal to $\delta_X$. Then we can estimate $c(X)$ as follows. 
\begin{equation}
    c(X) = \Big(\frac{0.144}{\delta_X}\Big)^2.
\end{equation}
For the previously studied datasets, this formula provides an estimate of $c \sim 0.33$. In our experiments, we found that this value works quite well; however, we found that sometimes adjusting this value (e.g., to $0.05$) provides better results, probably because the image representations computed by deep CNNs pretrained on ImageNet may not have been entirely accurate.

\section{Hyperbolic operations}\label{sec:hyp-neural}
Hyperbolic spaces are not vector spaces in a traditional sense; one cannot use standard operations as summation, multiplication, etc. To remedy this problem, one can utilize the formalism of  M\"obius gyrovector spaces allowing to generalize many standard operations to hyperbolic spaces. Recently proposed hyperbolic neural networks adopt this formalism to define the hyperbolic versions of feed-forward networks, multinomial logistic regression, and recurrent neural networks ~\cite{ganea2018hyperbolic}. 
In Appendix ~\ref{app:hypnetworks}, we discuss these networks and layers in detail, and in this section, we briefly summarize various operations available in the hyperbolic space. Similarly to the paper \cite{ganea2018hyperbolic}, we use an additional hyperparameter $c$ which modifies the curvature of Poincar\'e ball; it is then defined as \mbox{$\mathbb{D}_c^n = \{\mathbf{x} \in \mathbb{R}^n: c\|\mathbf{x}\|^2 <1, c \geq 0\}$}. The corresponding conformal factor now takes the form $\lambda_{\mathbf{x}}^c = \frac{2}{1 - c\|\mathbf{x}\|^2}$. In practice, the choice of $c$ allows one to balance between hyperbolic and Euclidean geometries, which is made precise by noting that with $c \to 0$, all the formulas discussed below take their usual Euclidean form. The following operations are the main building blocks of hyperbolic networks.

\paragraph{M\"obius addition.} 
For a pair $\mathbf{x}, \mathbf{y} \in \mathbb{D}_c^n$, the M\"obius addition is defined as follows:
\begin{equation}
    \mathbf{x} \oplus_c \mathbf{y} \coloneqq \frac{(1+2c\langle \mathbf{x}, \mathbf{y} \rangle + c\|\mathbf{y}\|^2) \mathbf{x}+ (1-c\|\mathbf{x}\|^2)\mathbf{y}}{1+2c\langle \mathbf{x}, \mathbf{y} \rangle + c^2 \|\mathbf{x}\|^2 \|\mathbf{y}\|^2}.
\end{equation}

\paragraph{Distance.} 
The induced distance function is defined as 
\begin{equation}
    d_c(\mathbf{x},\mathbf{y}) \coloneqq \frac{2}{\sqrt{c}} \mathrm{arctanh}(\sqrt{c}\|-\mathbf{x} \oplus_c \mathbf{y}\|).
\end{equation}
Note that with $c=1$ one recovers the geodesic distance \eqref{dist}, while with $c \to 0$ we obtain the Euclidean distance $\lim_{c \to 0} d_c(\mathbf{x},\mathbf{y})=2\|\mathbf{x}-\mathbf{y}\|.$

\paragraph{Exponential and logarithmic maps.} 
To perform operations in the hyperbolic space, one first needs to define a bijective map from $\mathbb{R}^n$ to $\mathbb{D}_c^n$ in order to map Euclidean vectors to the hyperbolic space, and vice versa.  The so-called exponential and (inverse to it) logarithmic map serves as such a bijection. 

The \emph{exponential} map $\exp_\mathbf{x}^c$ is a function from \mbox{$T_\mathbf{x} \mathbb{D}_c^n \cong \mathbb{R}^n$} to $\mathbb{D}_c^n$, which is given by
\begin{equation}\label{eq:exp}
    \exp_\mathbf{x}^c(\mathbf{v}) \coloneqq \mathbf{x} \oplus_ c \bigg(\tanh \bigg(\sqrt{c} \frac{\lambda_\mathbf{x}^c \|\mathbf{v}\|}{2} \bigg) \frac{\mathbf{v}}{\sqrt{c}\|\mathbf{v}\|}\bigg).
\end{equation}

The inverse \emph{logarithmic} map is defined as
\begin{equation}\label{eq:log}
    \log_\mathbf{x}^c(\mathbf{y}) \coloneqq  \frac{2}{\sqrt{c} \lambda_\mathbf{x}^c} \mathrm{arctanh}(\sqrt{c}\|-\mathbf{x} \oplus_c \mathbf{y}\|) \frac{-\mathbf{x}\oplus_c \mathbf{y}}{\|-\mathbf{x}\oplus_c \mathbf{y}\|}.
\end{equation}

In practice, we use the maps $\exp_\mathbf{0}^c$ and $\log_\mathbf{0}^c$ for a transition between the Euclidean and Poincar\'e ball representations of a vector.

\paragraph{Hyperbolic averaging.}
One important operation common in image processing is averaging of feature vectors, used, e.g., in prototypical networks for few--shot learning ~\cite{snell2017prototypical_net}. In the Euclidean setting this operation takes the form $(\mathbf{x}_1, \dots, \mathbf{x}_N) \to \frac{1}{N}\sum_i \mathbf{x}_i$. Extension of this operation to hyperbolic spaces is called the \emph{Einstein midpoint} and takes the most simple form in \emph{Klein} coordinates:
\begin{equation}\label{eq:hypave}
    \mathrm{HypAve}(\mathbf{x}_1, \dots, \mathbf{x}_N) = \sum_{i=1}^N \gamma_i \mathbf{x}_i / \sum_{i=1}^N \gamma_i,
\end{equation}
where $\gamma_i = \frac{1}{\sqrt{1 - c\|\mathbf{x}_i\|^2}}$ are the Lorentz factors. Recall from the discussion in Section \ref{sec:pball} that the Klein model is supported on the same space as the Poincar\'e ball; however, the same point has different coordinate representations in these models. Let $\mathbf{x}_{\mathbb{D}}$ and $\mathbf{x}_{\mathbb{K}}$ denote the coordinates of the same point in the Poincar\'e and Klein models correspondingly. Then the following transition formulas hold.
\begin{align}
\mathbf{x}_{\mathbb{D}} &= \frac{\mathbf{x}_{\mathbb{K}}}{1 + \sqrt{1 - c\| \mathbf{x}_{\mathbb{K}} \|^2}}, \\
\mathbf{x}_{\mathbb{K}} &= \frac{2\mathbf{x}_{\mathbb{D}}}{1 + c \|\mathbf{x}_{\mathbb{D}}\|^2}.
\end{align}
Thus, given points in the Poincar\'e ball, we can first map them to the Klein model, compute the average using Equation \eqref{eq:hypave}, and then move it back to the Poincar\'e model.

\paragraph{Numerical stability.}
While implementing most of the formulas described above is straightforward, we employ some tricks to make the training more stable. In particular, to ensure numerical stability, we perform clipping by norm after applying the exponential map, which constrains the norm not to exceed $\frac{1}{\sqrt{c}} (1 - 10^{-3})$.

\section{Experiments}\label{sec:experiments}

\begin{figure*}[htb!]
    \centering
    \includegraphics[width=0.95\linewidth]{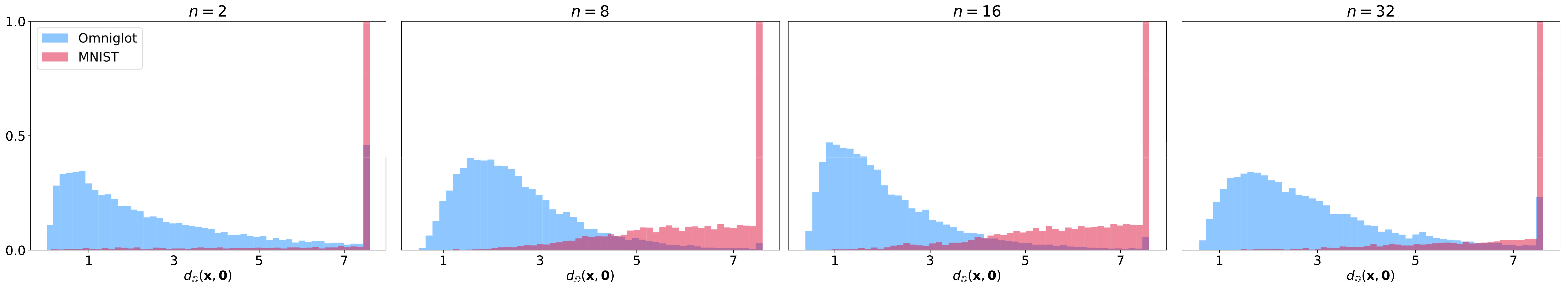}
    \caption{Distributions of the hyperbolic distance to the origin of the MNIST (red) and Omniglot (blue) datasets embedded into the Poincar\'e ball; parameter $n$ denotes embedding dimension of the model trained for MNIST classification. Most Omniglot instances can be easily identified as out-of-domain based on their distance to the origin.
     }
     \label{fig:dists_omni}
     \vspace{-5mm}
\end{figure*}

\paragraph{Experimental setup.}
We start with a toy experiment supporting our hypothesis that the distance to the center in Poincar\'e ball indicates a model uncertainty. To do so, we first train a classifier in hyperbolic space on the MNIST dataset ~\cite{lecun1998gradient} and evaluate it on the Omniglot dataset ~\cite{lake2013one}. We then investigate and compare the obtained distributions of distances to the origin of hyperbolic embeddings of the MNIST and Omniglot test sets. 

In our further experiments, we concentrate on the few-shot classification and person re-identification tasks. The experiments on the Omniglot dataset serve as a starting point, and then we move towards more complex datasets. Afterwards, we consider two datasets, namely: \textit{Mini}ImageNet ~\cite{ravi2016optimization} and Caltech-UCSD Birds-200-2011 (CUB) ~\cite{wah2011caltech}.
Finally, we provide the re-identification results for the two popular datasets: Market-1501 ~\cite{zheng2015scalable} and DukeMTMD ~\cite{ristani2016MTMC,zheng2017unlabeled}.
Further in this section, we provide a thorough description of each experiment. Our code is available at github\footnote{\href{https://github.com/leymir/hyperbolic-image-embeddings}{https://github.com/leymir/hyperbolic-image-embeddings}}.

\begin{table}[htb!]
\caption{Kolmogorov-Smirnov distances between the distributions of distance to the origin of the MNIST and Omniglot datasets embedded into the Poincar\'e ball with the hyperbolic classifier trained on MNIST, and between the distributions of $p_{\max}$ (maximum probablity predicted for a class) for the Euclidean classifier trained on MNIST and evaluated on the same sets.}
\label{tab:ks_omni}
\vspace{-5mm}
\begin{center}
  \begin{tabular}{lcccc}
    \toprule
  {} &  {$n=2$} & {$n=8$} & {$n=16$} & {$n=32$}\\
    \midrule
    $d_{\mathbb{D}}(\mathbf{x}, \mathbf{0})$ & $\mathbf{0.868}$ & $0.832$ & $\mathbf{0.853}$ & $\mathbf{0.859}$\\
    $p_{\max}(\mathbf{x})$ & $0.834$ & $\mathbf{0.835}$ & $0.840$ & $0.846$\\
    
    \bottomrule
  \end{tabular}
\end{center}
\vspace{-7mm}
\end{table}

\subsection{Distance to the origin as the measure of uncertainty}\label{subsec:distances}
In this subsection, we validate our hypothesis, which claims that if one trains a hyperbolic classifier, then the distance of the Poincar\'e ball embedding of an image to the origin can serve as a good measure of confidence of a model. We start by training a simple hyperbolic convolutional neural network on the MNIST dataset (we hypothesized that such a simple dataset contains a very basic hierarchy, roughly corresponding to visual ambiguity of images, as demonstrated by a trained network on Figure \ref{fig:teaser}). The output of the last hidden layer was mapped to the Poincar\'e ball using the exponential map \eqref{eq:exp} and was followed by the hyperbolic multi-linear regression (MLR) layer~\cite{ganea2018hyperbolic}.

After training the model to $\sim 99\%$ test accuracy, we evaluate it on the Omniglot dataset (by resizing its images to $28 \times 28$ and normalizing them to have the same background color as MNIST). We then evaluated the hyperbolic distance to the origin of embeddings produced by the network on both datasets. The closest Euclidean analogue to this approach would be comparing distributions of $p_{\max}$, maximum class probability predicted by the network. For the same range of dimensions, we train ordinary Euclidean classifiers on MNIST and compare these distributions for the same sets.
Our findings are summarized in Figure \ref{fig:dists_omni} and Table \ref{tab:ks_omni}. We observe that distances to the origin represent a better indicator of the dataset dissimilarity in three out of four cases.


We have visualized the learned MNIST and Omniglot embeddings in Figure \ref{fig:teaser}. We observe that more ``unclear'' images are located near the center, while the images that are easy to classify are located closer to the boundary.

\subsection{Few--shot classification}\label{subsec:omni}

We hypothesize that a certain class of problems --- namely the few-shot classification task can benefit from hyperbolic embeddings, due to the ability of hyperbolic space to accurately reflect even very complex hierarchical relations between data points. In principle, any metric learning approach can be modified to incorporate the hyperbolic embeddings. We decided to focus on the classical approach called prototypical networks (ProtoNets) introduced in ~\cite{snell2017prototypical_net}. This approach was picked because it is simple in general and simple to convert to hyperbolic geometry. ProtoNets use the so-called \emph{prototype representation} of a class, which is defined as a mean of the embedded support set of a class. Generalizing this concept to hyperbolic space, we substitute the Euclidean mean operation by $\mathrm{HypAve}$, defined earlier in \eqref{eq:hypave}. We show that Hyperbolic ProtoNets can achieve results competitive with many recent state-of-the-art models. Our main experiments are conducted on \textit{Mini}ImageNet and Caltech-UCSD Birds-200-2011 (CUB). Additional experiments on the Omniglot dataset, as well as the implementation details and hyperparameters, are provided in \Cref{app:details}. For a visualization of learned embeddings see \Cref{app:visualization}.

\paragraph{\textbf{\textit{Mini}}ImageNet.}
\textit{Mini}ImageNet dataset is the subset of ImageNet dataset ~\cite{russakovsky2015imagenet} that contains $100$ classes represented by $600$ examples per class. We use the following split provided in the paper ~\cite{ravi2016optimization}: the training dataset consists of $64$ classes, the validation dataset is represented by $16$ classes, and the remaining $20$ classes serve as the test dataset. We test the models on tasks for 1-shot and 5-shot classifications; the number of query points in each batch always equals to $15$. Similarly to \cite{snell2017prototypical_net}, the model is trained in the 30-shot regime for the 1-shot task and the 20-shot regime for the 1-shot task. We test our approach with two different backbone CNN models: a commonly used four-block CNN~\cite{snell2017prototypical_net,chen2018a} (denoted `4~Conv' in the table) and ResNet18~\cite{he2016deep}. To find the best values of hyperparameters, we used the grid search; see \Cref{app:details} for the complete list of values.

\begin{table}[t]
\caption{Few-shot classification accuracy results on \textit{Mini}ImageNet on 1-shot 5-way and 5-shot 5-way tasks. All accuracy results are reported with 95\% confidence intervals. 
}
\centering
 \scalebox{0.76}{
\begin{tabular}{l|c c c}
\Xhline{3\arrayrulewidth}
\textbf{Baselines}  & \textbf{Embedding Net} & \textbf{1-Shot 5-Way} & \textbf{5-Shot 5-Way} \\ 
\Xhline{2\arrayrulewidth}
MatchingNet~\cite{nips2016_vinyals2016matching_net}  & 4 Conv   & 43.56 $\pm$ 0.84\%   & 55.31 $\pm$ 0.73\%               \\ 
MAML~\cite{finn2017maml}      & 4 Conv            & 48.70 $\pm$ 1.84\%   & 63.11 $\pm$ 0.92\%             \\ 
RelationNet~\cite{sung2018relation_net} & 4 Conv           & 50.44 $\pm$ 0.82\%   & 65.32 $\pm$ 0.70\%            \\ 
REPTILE~\cite{nichol2018reptile}      &   4 Conv       & 49.97 $\pm$ 0.32\%   & 65.99 $\pm$ 0.58\%                  \\ 
ProtoNet~\cite{snell2017prototypical_net} &     4 Conv         & 49.42 $\pm$ 0.78\%   & 68.20 $\pm$ 0.66\%             \\ 
Baseline*~\cite{chen2018a}  &       4 Conv        & 41.08 $\pm$ 0.70\%         & 54.50 $\pm$ 0.66\%              \\ 
Spot\&learn~\cite{chu_cvpr2019_spot_and_learn}  &       4 Conv        & 51.03 $\pm$ 0.78\%         & 67.96 $\pm$ 0.71\%                 \\ 
DN4~\cite{li_cvpr2019_revisiting}  &       4 Conv        & 51.24 $\pm$ 0.74\%         & 71.02 $\pm$ 0.64\%                  \\ \hline

\textbf{Hyperbolic ProtoNet} & 4 Conv & \textbf{54.43 $\pm$ 0.20\%} & \textbf{72.67 $\pm$ 0.15\%} \\

\hline

SNAIL~\cite{mishra2017snail}    & ResNet12
& 55.71 $\pm$ 0.99\%   & 68.88 $\pm$ 0.92\%                \\ 
ProtoNet$^{+}$~\cite{snell2017prototypical_net} & ResNet12
& 56.50 $\pm$ 0.40\% & 74.2 $\pm$ 0.20\%    \\

CAML~\cite{jiang2018learning_caml}      &  ResNet12      & 59.23 $\pm$ 0.99\%   & 72.35 $\pm$ 0.71\%            \\ 

TPN~\cite{liu2018learning_TPN}        &   ResNet12
& 59.46\%         & 75.65\%           \\ 
MTL~\cite{sun2019meta_mtl}        & ResNet12
& 61.20 $\pm$ 1.8\%   & 75.50 $\pm$ 0.8\%                \\
DN4~\cite{li_cvpr2019_revisiting}        & ResNet12
& 54.37 $\pm$ 0.36\%   & 74.44 $\pm$ 0.29\%               \\
TADAM~\cite{oreshkin2018tadam}      &    ResNet12
& 58.50\%         & 76.70\%                 \\

Qiao-WRN~\cite{qiao2018few_shot_activations}   &    Wide-ResNet28
& 59.60 $\pm$ 0.41\%   & 73.74 $\pm$ 0.19\%            \\ 
LEO~\cite{rusu2018LEO}        & Wide-ResNet28
& \textbf{61.76 $\pm$ 0.08\%}   & \textbf{77.59 $\pm$ 0.12\%}                \\
Dis. k-shot~\cite{bauer2017discriminative_k_shot} & ResNet34  & 56.30 $\pm$ 0.40\%   & 73.90 $\pm$ 0.30\%       \\ 
Self-Jig(SVM)~\cite{chen2019aug_few_shot}     &   ResNet50
& 58.80 $\pm$ 1.36\%   & 76.71 $\pm$ 0.72\%             \\ \hline
\textbf{Hyperbolic ProtoNet} & ResNet18 & 59.47 $\pm$ 0.20\%& 76.84 $\pm$ 0.14\%\\
\Xhline{3\arrayrulewidth}
\end{tabular}
}
\vspace{-5mm}
\label{tab:min_final}
\end{table}

Table~\ref{tab:min_final} illustrates the obtained results on the \textit{Mini}ImageNet dataset (alongside other results in the literature). Interestingly, \textbf{Hyperbolic ProtoNet} significantly improves accuracy as compared to the standard ProtoNet, especially in the one-shot setting. We observe that the obtained accuracy values, in many cases, exceed the results obtained by more advanced methods, sometimes even in the case of architecture of larger capacity. This partly confirms our hypothesis that hyperbolic geometry indeed allows for more accurate embeddings in the few--shot setting.


\paragraph{Caltech-UCSD Birds.}
The CUB dataset consists of $11,788$ images of $200$ bird species and was designed for fine-grained classification. We use the split introduced in ~\cite{triantafillou2017few}: $100$ classes out of $200$ were used for training, $50$ for validation and $50$ for testing. Due to the relative simplicity of the dataset, we consider only the 4-Conv backbone and do not modify the training shot values as was done for the \textit{Mini}ImageNet case. The full list of hyperparameters is provided in \Cref{app:details}. 

Our findings are summarized in Table \ref{tab:cub_final}. Interestingly, for this dataset, the hyperbolic version of ProtoNet significantly outperforms its Euclidean counterpart (by more than 10\% in both settings), and outperforms many other algorithms.
\begin{table}[!ht]
\caption{Few-shot classification accuracy results on CUB dataset~\cite{WahCUB_200_2011} on 1-shot 5-way task, 5-shot 5-way task. All accuracy results are reported with 95\% confidence intervals.
For each task, the best-performing method is highlighted.}
\centering
\scalebox{0.76}{
\begin{tabular}{l|c c c}
\Xhline{3\arrayrulewidth}
\textbf{Baselines}      & Embedding Net & \textbf{1-Shot 5-Way} & \textbf{5-Shot 5-Way} \\ 
\Xhline{2\arrayrulewidth}
MatchingNet~\cite{nips2016_vinyals2016matching_net} & 4 Conv & 61.16 $\pm$ 0.89   & 72.86 $\pm$ 0.70   \\ 
MAML~\cite{finn2017maml}    &  4 Conv  & 55.92 $\pm$ 0.95\%   & 72.09 $\pm$ 0.76\%   \\ 
ProtoNet~\cite{snell2017prototypical_net} & 4 Conv  & 51.31 $\pm$ 0.91\%   & 70.77 $\pm$ 0.69\%   \\ 
MACO~\cite{hilliard2018few_conditional_embedding_maco}   &  4 Conv   & 60.76\%        & 74.96\%        \\ 
RelationNet~\cite{sung2018relation_net} & 4 Conv & 62.45 $\pm$ 0.98\%   & 76.11 $\pm$ 0.69\%   \\ 
Baseline++~\cite{chen2018a} & 4 Conv & 60.53 $\pm$ 0.83\%   & 79.34 $\pm$ 0.61\%   \\ 
DN4-DA~\cite{li_cvpr2019_revisiting} & 4 Conv & 53.15 $\pm$ 0.84\%   & 81.90 $\pm$ 0.60\%   \\ \hline
\textbf{Hyperbolic ProtoNet} & 4 Conv & \textbf{64.02 $\pm$ 0.24\%} & \textbf{82.53 $\pm$ 0.14\%}\\

\Xhline{3\arrayrulewidth}

\end{tabular}
}
\label{tab:cub_final}
\end{table}

\subsection{Person re-identification}\label{subsec:reid}
\begin{table}[t]
\centering
\caption{Person re-identification results for Market-1501 and DukeMTMC-reID for the classification baseline (\textit{Euclidean}) and its hyperbolic counterpart (\textit{Hyperbolic}). (See \ref{subsec:reid} for the details). The results are shown for the three embedding dimensionalities and for two different learning rate schedules. For each dataset and each embedding dimensionality value, the best results are bold, they are all given by the hyperbolic version of classification (either by the schedule \textit{sch\#1} or \textit{sch\#2}). The second-best results are underlined. }
\label{tab:reid}
\scalebox{0.74}{
\begin{tabular}{cccccccccc}
\Xhline{3\arrayrulewidth}
                                & \multicolumn{4}{c}{Market-1501}                                                                     & \multicolumn{4}{c}{DukeMTMC-reID}                                                                            \\
\midrule
                                & \multicolumn{2}{c}{Euclidean}                           & \multicolumn{2}{c}{Hyperbolic}                          & \multicolumn{2}{c}{Euclidean}                           & \multicolumn{2}{c}{Hyperbolic}                          \\

                \multicolumn{1}{c}{dim, lr schedule} & \multicolumn{1}{c}{r1} & \multicolumn{1}{c}{mAP} & \multicolumn{1}{c}{r1} & \multicolumn{1}{c}{mAP} & \multicolumn{1}{c}{r1} & \multicolumn{1}{c}{mAP} & \multicolumn{1}{c}{r1} & \multicolumn{1}{c}{mAP} \\
\midrule
32, sch\#1                          & {\ul 71.4}             & {\ul 49.7}              & 69.8                   & 45.9                    & 56.1                   & 35.6                    & 56.5                   & 34.9                    \\
                      32, sch\#2                          & 68.0                   & 43.4                    & \textbf{75.9}          & \textbf{51.9}           & {\ul 57.2}             & {\ul 35.7}              & \textbf{62.2}          & \textbf{39.1}           \\
                     \midrule
 64, sch\#1                          & 80.3                   & 60.3                    & {\ul 83.1}             & {\ul 60.1}              & 69.9                   & 48.5                    & \textbf{70.8}          & \textbf{48.6}           \\
                      64, sch\#2                          & 80.5                   & 57.8                    & \textbf{84.4}          & \textbf{62.7}           & 68.3                   & 45.5                    & {\ul 70.7}             & {\ul 48.6}              \\
                     \midrule
 128, sch\#1                          & 86.0                   & 67.3                    & \textbf{87.8}          & \textbf{68.4}           & {\ul 74.1}             & {\ul 53.3}              & \textbf{76.5}          & \textbf{55.4}           \\
                      128, sch\#2                          & {\ul 86.5}             & {\ul 68.5}              & 86.4                   & 66.2                    & 71.5                   & 51.5                    & 74.0                   & 52.2        \\           
\Xhline{3\arrayrulewidth}
\end{tabular}
}
\vspace{-5mm}
\end{table}

The DukeMTMC-reID dataset~\cite{ristani2016MTMC,zheng2017unlabeled} contains $16,522$ training images of $702$ identities, $2,228$ query images of $702$ identities and $17,661$ gallery images. The Market1501 dataset~\cite{zheng2015scalable} contains $12,936$ training images of $751$ identities, $3,368$ queries of $750$ identities and $15,913$ gallery images respectively. We report Rank1 of the Cumulative matching Characteristic Curve and Mean Average Precision for both datasets. The results (Table~\ref{tab:reid}) are reported after the $300$ training epochs. The experiments were performed with the ResNet50 backbone, and two different learning rate schedulers (see Appendix~\ref{app:details} for more details). The hyperbolic version generally performs better than the Euclidean baseline, with the advantage being bigger for smaller dimensionality.

\section{Discussion and conclusion}\label{sec:conclusion}
We have investigated the use of hyperbolic spaces for image embeddings. The models that we have considered use Euclidean operations in most layers, and use the exponential map to move from the Euclidean to hyperbolic spaces at the end of the network (akin to the normalization layers that are used to map from the Euclidean space to Euclidean spheres). The approach that we investigate here is thus compatible with existing backbone networks trained in Euclidean geometry.

At the same time, we have shown that across a number of tasks, in particular in the few-shot image classification, learning hyperbolic embeddings can result in a substantial boost in accuracy. We speculate that the negative curvature of the hyperbolic spaces allows for embeddings that are better conforming to the intrinsic geometry of at least some image manifolds with their hierarchical structure.

Future work may include several potential modifications of the approach. We have observed that the benefit of hyperbolic embeddings may be substantially bigger in some tasks and datasets than in others. A better understanding of when and why the use of hyperbolic geometry is warranted is therefore needed. Finally, we note that while all hyperbolic geometry models are equivalent in the continuous setting, fixed-precision arithmetic used in real computers breaks this equivalence. In practice, we observed that care should be taken about numeric precision effects. Using other models of hyperbolic geometry may result in a more favourable floating point performance.

\section*{Acknowledgements} This work was funded by the Ministry of Science and Education of Russian Federation as a part of Mega Grant Research Project 14.756.31.000.

\bibliographystyle{ieee_fullname}
\bibliography{main}
\clearpage
\appendix
\section{Hyperbolic Neural Networks}{\label{app:hypnetworks}}

\paragraph{Linear layer.}
Assume we have a standard (Euclidean) linear layer $\mathbf{x} \to \mathrm{M}\mathbf{x} + \mathbf{b}$. In order to generalize it, one needs to define the M\"obius matrix by vector product:
\begin{equation}
    \mathrm{M}^{\otimes_c} (\mathbf{x}) \coloneqq \frac{1}{\sqrt{c}} \tanh \bigg(\frac{\|\mathrm{M}\mathbf{x}\|}{\|\mathbf{x}\|}\mathrm{arctanh}(\sqrt{c} \|\mathbf{x}\|)\bigg)\frac{\mathrm{M}\mathbf{x}}{\|\mathrm{M}\mathbf{x}\|},
\end{equation}
if $\mathrm{M}\mathbf{x} \neq \mathbf{0}$, and  $\mathrm{M}^{\otimes_c} (\mathbf{x}) \coloneqq \mathbf{0}$ otherwise.
Finally, for a bias vector $\mathbf{b} \in \mathbb{D}^n_c$ the operation underlying the hyperbolic linear layer is then given by $\mathrm{M}^{\otimes_c} (\mathbf{x}) \oplus_c \mathbf{b}$.

\paragraph{Concatenation of input vectors.}
In several architectures (e.g., in siamese networks), it is needed to concatenate two vectors; such operation is obvious in Euclidean space. However, straightforward concatenation of two vectors from hyperbolic space does not necessarily remain in hyperbolic space. Thus, we have to use a generalized version of the concatenation operation, which is then defined in the following manner. For $\mathbf{x} \in \mathbb{D}^{n_1}_c$, $\mathbf{y} \in \mathbb{D}^{n_2}_c$ we define the mapping \mbox{$\mathrm{Concat}: \mathbb{D}^{n_1}_c \times \mathbb{D}^{n_2}_c \to \mathbb{D}^{n_3}_c$} as follows.
\begin{equation}{\label{eq:concat}}
    \mathrm{Concat}(\mathbf{x}, \mathbf{y}) = \mathrm{M}_1^{\otimes_c} \mathbf{x} \oplus_c \mathrm{M}_2^{\otimes_c} \mathbf{y},
\end{equation}
where $\mathrm{M}_1$ and $\mathrm{M}_2$ are trainable matrices of sizes $n_3 \times n_1 $ and $n_3 \times n_2$ correspondingly. The motivation for this definition is simple: usually, the Euclidean concatenation layer is followed by a linear map, which when written explicitly takes the (Euclidean) form of Equation \eqref{eq:concat}.

\paragraph{Multiclass logistic regression (MLR).} 
In our experiments, to perform the multiclass classification, we take advantage of the generalization of multiclass logistic regression to hyperbolic spaces. The idea of this generalization is based on the observation that in Euclidean space logits can be represented as the distances to certain \emph{hyperplanes}, where each hyperplane can be specified with a point of origin and a normal vector. The same construction can be used in the Poincar\'e ball after a suitable analogue for hyperplanes is introduced. Given $\mathbf{p} \in \mathbb{D}_c^n$ and $\mathbf{a} \in T_{\mathbf{p}}\mathbb{D}_c^n \setminus \{\mathbf{0}\}$, such an analogue would be the union of all geodesics passing through $\mathbf{p}$ and orthogonal to $\mathbf{a}$.


The resulting formula for hyperbolic MLR for $K$ classes is written below; here $\mathbf{p}_k \in \mathbb{D}_c^n$ and $\mathbf{a}_k \in T_{\mathbf{p}_k}\mathbb{D}_c^n \setminus \{\mathbf{0}\}$ are learnable parameters.

\begin{equation*}
\begin{split}
    p(y &=k | \mathbf{x}) \propto  \\
    & \exp \bigg( \frac{\lambda_{\mathbf{p}_k}^c\|\mathbf{a}_k\|}{\sqrt{c}}\mathrm{arcsinh}\bigg(\frac{2\sqrt{c}\langle -\mathbf{p}_k \oplus_c \mathbf{x}, \mathbf{a}_k \rangle}{(1-c\|-\mathbf{p}_k \oplus_c \mathbf{x}\|^2)\|\mathbf{a}_k\|}\bigg)\bigg).
\end{split}
\end{equation*}

For a more thorough discussion of hyperbolic neural networks, we refer the reader to the paper ~\cite{ganea2018hyperbolic}. 

\section{Experiment details}{\label{app:details}}
\paragraph{Omniglot.}
As a baseline model, we consider the prototype network (ProtoNet). Each convolutional block consists of $3 \times 3$ convolutional layer followed by batch normalization, ReLU nonlinearity and $2 \times 2$ max-pooling layer. The number of filters in the last convolutional layer corresponds to the value of the embedding dimension, for which we choose $64$. The hyperbolic model differs from the baseline in the following aspects. First, the output of the last convolutional block is embedded into the Poincar\'e ball of dimension $64$ using the exponential map. 
Results are presented in Table \ref{tab:omni_proto}. We can see that in some scenarios, in particular for one-shot learning, hyperbolic embeddings are more beneficial, while in other cases, results are slightly worse. The relative simplicity of this dataset may explain why we have not observed a significant benefit of hyperbolic embeddings. We further test our approach on more advanced datasets.
\begin{table}[htb!]
\caption{Few-shot classification accuracies on Omniglot. In order to obtain Hyperbolic ProtoNet, we augment the standard ProtoNet with a mapping to the Poincar\'e ball, use hyperbolic distance as the distance function, and as the averaging operator we use the $\mathrm{HypAve}$ operator defined by Equation \eqref{eq:hypave}.}
\label{tab:omni_proto}
\begin{center}
  \begin{tabular}{lcc}
\toprule
  {} & {ProtoNet} & {Hyperbolic ProtoNet} \\
    \midrule
    {$1$-shot $5$-way} & $98.2$ & $\mathbf{99.0}$ \\
    {$5$-shot $5$-way} & $99.4$ & $99.4$ \\
    {$1$-shot $20$-way} & $95.8$ & $\mathbf{95.9}$ \\
    {$5$-shot $20$-way} & $\mathbf{98.6}$ & $98.15$ \\
    \bottomrule
  \end{tabular}
\end{center}
\vspace{-7mm}
\end{table}

\paragraph{\textit{mini}ImageNet.}
We performed the experiments with two different backbones, namely the previously discussed 4-Conv model and ResNet18. For the former, embedding dim was set to 1024 and for the latter to 512. For the one-shot setting both models were trained for $200$ epochs with Adam optimizer, learning rate being $5 \cdot 10^{-3}$ and step learning rate decay with the factor of $0.5$ and step size being $80$ epochs. For the 4-Conv model we used $c=0.01$ and for ResNet18 we used $c=0.001$.
For 4-Conv in the five-shot setting we used the same hyperparameters except for $c=0.005$ and learning rate decay step being $60$ epochs. For ResNet18 we additionally changed learning rate to $10^{-3}$ and step size to $40$.




\paragraph{Caltech-UCSD Birds.}
For these experiments we used the same 4-Conv architecture with the embedding dimensionality being $512$. For the one-shot task, we used learning rate $10^{-3}$, $c=0.05$, learning rate step being $50$ epochs and decay rate of $0.8$. For the five-shot task, we used learning rate $10^{-3}$, $c=0.01$, learning rate step of $40$ and decay rate of $0.8$.


\paragraph{Person re-identification.}
We use ResNet50 ~\cite{he2016deep} architecture with one fully connected embedding layer following the global average pooling. Three embedding dimensionalities are used in our experiments: $32$, $64$ and $128$.
For the baseline experiments, we add the additional classification linear layer, followed by the cross-entropy loss.
For the hyperbolic version of the experiments, we map the descriptors to the Poincar\'e ball and apply multiclass logistic regression as described in Section \ref{sec:hyp-neural}.
We found that in both cases the results are very sensitive to the learning rate schedules. We tried four schedules for learning  $32$-dimensional descriptors for both baseline and hyperbolic versions. The two best performing schedules were applied for the $64$ and $128$-dimensional descriptors. 
In these experiments, we also found that smaller $c$ values give better results. We therefore have set $c$ to $10^{-5}$. Based on the discussion in \ref{sec:hyp-neural}, our hyperbolic setting is quite close to Euclidean.
The results are compiled in Table \ref{tab:reid}. 
We set starting learning rates to $3 \cdot 10^{-4}$ and $6 \cdot 10^{-4}$ for $sch\#1$ and $sch\#2$ correspondingly and multiply them by $0.1$ after each of the epochs $200$ and $270$.
\section{Visualizations}{\label{app:visualization}}
\begin{figure*}[htb!]
    \centering
    \includegraphics[width=0.9\linewidth]{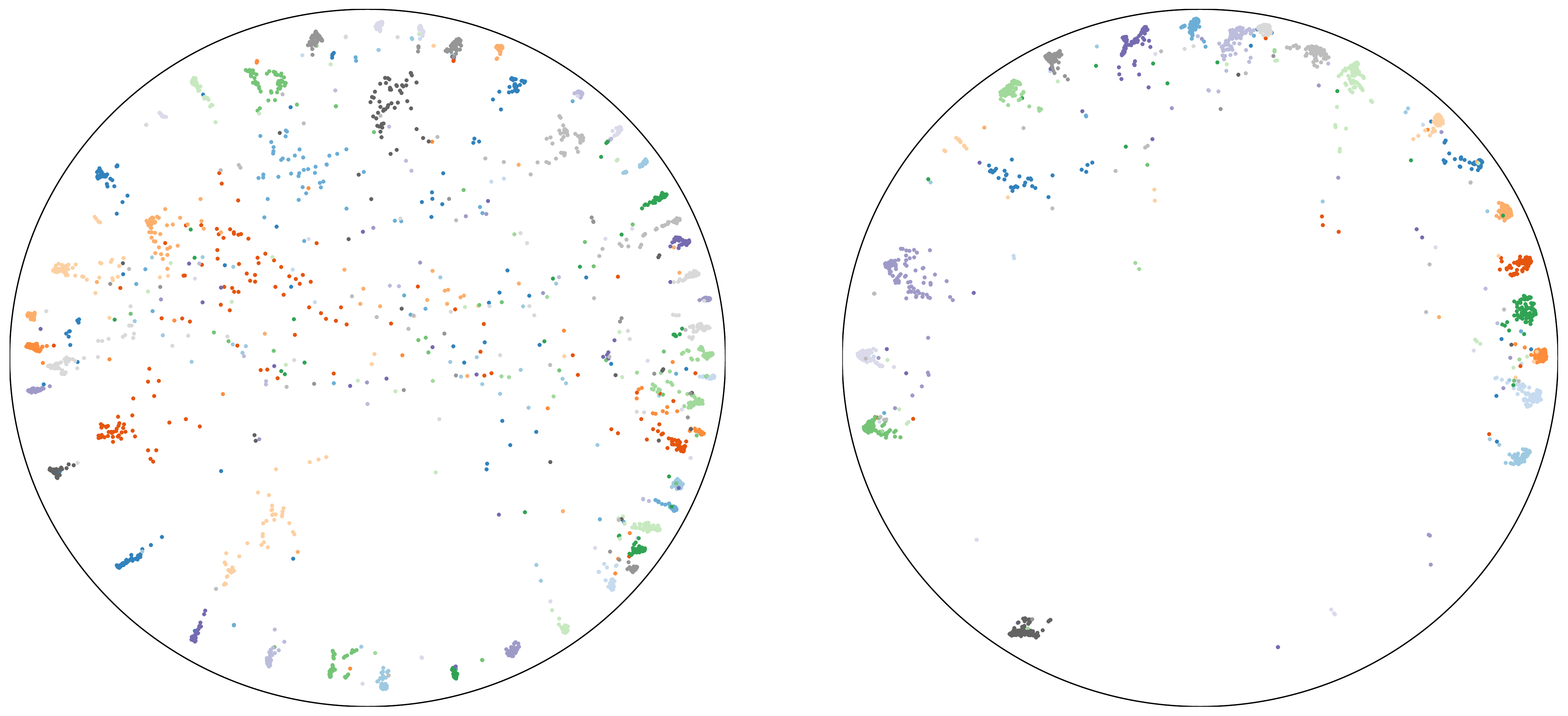}
    \caption{A visualization of the hyperbolic embeddings learned for the few--shot task. \textbf{Left}: 5-shot task on CUB. \textbf{Right}: 5-shot task on \emph{Mini}ImageNet. The two-dimensional projection was computed with the UMAP algorithm \cite{mcinnes2018umap}.
    }
    \label{fig:embeds}
\end{figure*}
For the visual inspection of embeddings we computed projections of high dimensional embeddings obtained from the trained few--shot models with the (hyperbolic) UMAP algorithm \cite{mcinnes2018umap} (see \Cref{fig:embeds}). We observe that different classes are neatly positioned near the boundary of the circle and are well separated.
\end{document}